\documentclass[11pt]{article}
\usepackage{acl}
\usepackage{latexsym}
\usepackage[T1]{fontenc}
\usepackage[utf8]{inputenc}
\usepackage{times}
\usepackage{microtype}
\usepackage{graphicx}
\usepackage{booktabs}
\usepackage{amsmath}
\usepackage{amssymb}
\usepackage{placeins}
\usepackage{tikz}
\title{When Should LLM Agents Forget? Memory Filtering Under Query-Critical Resolver Errors}

\author{Samarth Vinayaka \\
  University of Florida \\
  \texttt{samarthvinayaka@ufl.edu}}

\begin{document}
\maketitle

\begin{abstract}
LLM memory systems often delete facts classified as stale. We show that this hard-filtering step can reduce downstream accuracy when resolver errors delete query-critical facts. Across three models, a controlled sensitivity intervention moves hard-drop from gains of $+0.186$ to $+0.230$ at perfect resolution to losses of $-0.176$ to $-0.181$ at the lowest tested reliability. Paired targeted interventions show that false-drop causes significantly greater degradation than false-keep, while a non-deleting baseline can still fail when incorrect labels steer generation. On a frozen 102/102 calibration/test split, a simple deployment-matched gate reaches 0.655 accuracy versus 0.641 for always hard-drop and 0.662 for test-informed oracle selection. Global-accuracy crossovers require gold labels and are model- and task-specific diagnostics, not deployment thresholds. Deletion should instead be treated as cost-sensitive: validate query-critical false-drop risk on matched data, or default to label-free retention when such validation is unavailable.
\end{abstract}

\section{Introduction}

LLM agents deployed over long sessions or across many interactions accumulate memory: facts about a user or environment that were true when written but may not remain true. A user who says ``I live in Boston'' this month may say ``I moved to Austin'' next month; a project deadline expires; a stated preference gets walked back and later reasserted. An agent that treats all remembered facts as equally current will, sooner or later, act on stale information. We show that resolution accuracy and downstream behavioral accuracy can dissociate in \emph{both} directions: a resolver can be perfect and the agent can still answer incorrectly, and an assembly strategy that helps under a reliable resolver can actively harm under an unreliable one.

A substantial body of work addresses this by improving \emph{resolution}: classifying each memory as current, superseded, or expired, via a write-time consolidation step, a retrieval-time re-ranking step, or an LLM-based conflict-resolution judgment. This line of work often implicitly assumes that better resolution accuracy translates fairly directly into better downstream behavior.

We show this assumption does not hold in general. A resolver can be perfect and an agent can still answer incorrectly, if the packaging of that resolved information does not make the correction usable by the generator. Figure~\ref{fig:example} illustrates the point concretely: identical, correctly-resolved facts produce different generator outputs depending purely on how they are assembled into context.

\begin{figure}[!htb]
\centering
\small
\begin{tabular}{p{0.45\textwidth}}
\toprule
\textbf{Memory chain} (query: \textit{``Where does the user live?''})\\
\quad Jan 1: ``I live in Boston.''\\
\quad Feb 1: ``I moved to Austin.''\\
\midrule
\textbf{Flat context} (concatenated, validity ignored):\\
\quad\textit{``The user lives in Boston. The user moved to Austin.''}\\
\quad Generator output: \textbf{``Boston''} -- incorrect\\
\midrule
\textbf{Structured context} (validity-typed, non-current dropped):\\
\quad\textit{``CURRENT: The user moved to Austin.''}\\
\quad Generator output: \textbf{``Austin''} -- correct\\
\bottomrule
\end{tabular}
\caption{A worked example: both contexts are built from the identical, correctly-resolved memory chain. Only the assembly format differs, and it changes whether the generator uses the correction.}
\label{fig:example}
\end{figure}

We formalize this by decomposing the pipeline into three independently manipulable stages -- \textbf{R}esolver, \textbf{A}ssembly, \textbf{G}enerator -- and holding the resolver and generator fixed while varying only assembly format. Prior work often couples resolution and assembly into a single system, making it difficult to attribute a behavioral improvement to better resolution versus better packaging of the same resolution; our design isolates assembly as an independently manipulated variable.

Our central empirical contributions are:

\begin{enumerate}
\item \textbf{A resolver-error sensitivity test.} Corrupting oracle labels at a known rate moves AssemblyGain through zero on all three models; targeted interventions show that dropping a query-critical current fact is substantially more damaging than retaining a stale one (\S\ref{sec:reliability}).
\item \textbf{Policy alternatives under resolver error.} Soft retention distinguishes deletion loss from label-induced steering but is not automatically safe; a frozen calibration-selected hard-drop/flat gate avoids most oracle regret across held-out corruption regimes (\S\ref{sec:reliability}).
\item \textbf{Oracle and published-benchmark diagnostics.} Oracle and format ablations isolate the benefit of filtering when labels are trustworthy; a tuned diagnostic on the published MemoryAgentBench benchmark exhibits the predicted extraction--deletion mechanism but is not presented as held-out external validation (\S\ref{sec:headline}--\S\ref{sec:mab}).
\end{enumerate}

The benefit of validity-aware assembly is \emph{mediated} by upstream resolver errors: filtering improves behavior when labels are accurate and can harm it when errors remove needed evidence. Structured, content-dropping assembly is therefore not strictly dominant over flat concatenation. The operational quantity is not global label accuracy alone, but the cost and incidence of query-critical false drops.

\section{Related Work}
\label{sec:related}

We position this work as orthogonal to memory-\emph{system} design: we do not propose a new memory store, retrieval mechanism, or consolidation algorithm. We isolate and characterize one interface -- how already-resolved validity information is packaged into a generator's context -- that existing systems bundle with their own resolution or retrieval mechanism, making it hard to attribute their reported gains specifically to packaging.

\textbf{Agent memory systems.} MemGPT-style paging \citep{packer2023memgpt} moves content between working and external context via LLM function calls, using ``in context'' as a proxy for relevance and recency rather than validity. Mem0 \citep{mem0-2025} performs write-time classification of each new fact against existing memories (add/update/delete/no-op). A-Mem \citep{amem2025} organizes memory via dynamic, Zettelkasten-style linking, optimizing retrieval relevance. Generative Agents \citep{park2023generative} and Reflexion \citep{shinn2023reflexion} both treat memory as an evolving textual or episodic record, without an explicit validity-typed resolution step. To our knowledge, none of these vary assembly format independently of their own retrieval or resolution mechanism, so a behavioral gain due to better resolution cannot be separated from one due to how resolved content is packaged.

\textbf{Sanity check against an end-to-end memory system.}
\label{sec:mem0sanity}
We ran Mem0 OSS v2.0.11 end-to-end on an $n=102$ \textsc{ValidityMem-530} subset; the version was recovered from the unchanged environment's installed-package metadata and timestamped before output creation. Mem0 used Claude Haiku 4.5 internally, FastEmbed with \texttt{BAAI/bge-small-en-v1.5}, and local Qdrant. Under the frozen grader, Mem0 reaches 0.480 versus subset-matched oracle-flat 0.578 and oracle-hard-drop 0.745. By scenario it scores 0.941 on \texttt{historical\_only}, 0.588 on \texttt{superseded\_monotonic}, 0.353 on \texttt{expired\_interval} and \texttt{current}, 0.412 on \texttt{implicit\_conflict}, and 0.235 on \texttt{non\_monotonic\_reversion}. We retain it as a sanity check rather than a controlled system comparison because retrieval, resolution, and assembly remain coupled.

\textbf{Context compression.} LongLLMLingua \citep{longllmlingua2023} performs query-conditioned, perplexity-based compression that is validity-agnostic by design -- it can compress away a needed correction as readily as anything else. Our \texttt{structured\_budget} condition (Appendix~\ref{sec:budget}) studies validity-\emph{aware} budget selection specifically, and shows that even a validity-aware selector can fail by dropping the fact that matters under a tight budget.

\textbf{Benchmarks.} MemoryAgentBench \citep{mab2025} evaluates four memory competencies -- accurate retrieval, test-time learning, long-range understanding, and selective forgetting -- via reconstructed and newly-introduced datasets including FactConsolidation, which we reuse as a published-benchmark diagnostic (\S\ref{sec:mab}). It reports end-task accuracy without a resolver-versus-assembly decomposition. We call our controlled diagnostic \textsc{ValidityMem-530} to avoid collision with the previously published FreshBench temporal-generalization benchmark \citep{zhu2025freshbench}.

\textbf{Adjacent memory-conflict and consolidation work.} MemConflict evaluates dynamic, static, and conditional memory conflicts through black-box answers and white-box retrieval/ranking analysis \citep{tao2026memconflict}. Selective QA over conflicting multi-source personal memory explicitly separates evidence availability from conflict resolution and abstention \citep{yang2026selectiveqa}. StateFuse directly contrasts conflict-preserving and collapsed memory surfaces \citep{volkov2026statefuse}. These studies establish that conflict structure and preservation matter; our complementary design holds candidate evidence fixed, manipulates packaging independently, and causally sweeps resolver errors until the behavioral effect reverses. Concurrent work also compares a resolver with a deterministic max-serial baseline on MAB FactConsolidation \citep{deterministicrecipe2026}; STALE \citep{stale2026} benchmarks resolution without a decomposed assembly factor; and Memanto \citep{memanto2026}, TiMem \citep{timem2026}, and Control-Plane Placement \citep{controlplane2026} study end-to-end typed memory or the placement of forgetting logic rather than an independently randomized packaging policy.

\emph{What distinguishes this work}: (1) an experimental design that holds resolution quality fixed and varies assembly policy; (2) interventional evidence that a content-dropping policy's benefit is conditional on upstream reliability and can reverse sign; and (3) separate reporting of resolution, query-critical false drops, evidence retention, and downstream behavior rather than one end-task number.

\section{Problem Formulation}
\label{sec:formulation}

Let $M = \{m_1, \ldots, m_k\}$ be the set of candidate memories relevant to a query $q$ issued at time $t$. Each memory has a ground-truth validity state $V(m)$, unknown to the system at time $t$, taking one of five values: \texttt{current}, \texttt{superseded}, \texttt{expired}, \texttt{historical}, or \texttt{conflicting}. We take $M$ as given, selected by an upstream \emph{retrieval} step outside our scope; an end-to-end system's retrieval, resolution, and assembly stages can confound one another (\S\ref{sec:mem0sanity} examines this directly), which is exactly why we hold $M$ fixed and vary only the stages below.

A \textbf{Resolver} $R: (M, t) \to \hat{V}$ produces a predicted validity label for each memory. An \textbf{Assembly} function $A: (M, \hat{V}, q, t) \to c$ applies a query- and time-dependent retention and presentation policy, producing a context string $c$ passed to a \textbf{Generator} $G: (c, q) \to a$. Query dependence is necessary: for example, a historical memory is retained for a historically scoped query but not treated as current evidence for a present-state query.

We report two metrics, kept explicitly separate rather than collapsed into one end-task number, plus two derived quantities used throughout:
\[
\text{ResolutionAcc} = \Pr[\hat{V}(m) = V(m)]
\]
computed over every candidate memory (not just whether the single current fact was identified -- a resolver that labels everything \texttt{current} would score perfectly on a recall-only variant of this metric without resolving anything). ResolutionAcc requires gold labels and is therefore a retrospective experimental diagnostic, not an observable deployment gate. We also report
\begin{multline*}
\text{BehavioralAcc} = {}\\
\Pr[G(A(M,R(M,t),q,t),q)=\text{gold}(q)]
\end{multline*}
the standard downstream task accuracy. Because these are measured at different levels (memory-level vs.\ query-level), we do not report their difference as a single scalar; we instead report:
\begin{multline*}
\text{OracleShortfall} = 1 - {}\\
\text{BehavioralAcc}(A_{\text{flat}}\mid R_{\text{oracle}})
\end{multline*}
how far even a perfectly-resolved, flat-assembled system falls short of ceiling, and, our central quantity of interest,
\begin{multline*}
\text{AssemblyGain} = \text{BehavioralAcc}(A_{\text{hard}}) \\
{} - \text{BehavioralAcc}(A_{\text{flat}})
\end{multline*}
computed within a fixed resolver tier, isolating the effect of assembly format under controlled resolver and generator settings.

\textbf{Experimental design.} We cross four resolver tiers with assembly modes, holding the generator fixed, to isolate assembly as an independently manipulated variable:

\emph{Resolver tiers}: \texttt{no\_labels} (a floor baseline with no validity signal), \texttt{oracle} (gold labels), \texttt{deterministic} (rule-based temporal ordering), and \texttt{hybrid} (deterministic rules with LLM adjudication escalation for ambiguous cases). \S\ref{sec:reliability} additionally sweeps a fifth, continuous condition: oracle labels corrupted at a controlled rate $p$.

\emph{Assembly modes}: \texttt{flat} (concatenate all memories, ignoring validity -- by construction invariant to resolver tier), \texttt{soft-retain} (retain every memory while labeling and reordering predicted current/non-current evidence), \texttt{hard-drop} (the legacy \texttt{structured} mode: retain predicted current or uncertain items, retain historical items only for historical queries, and drop predicted superseded/expired items), \texttt{structured\_budget} (hard-drop under a token budget, Appendix~\ref{sec:budget}), and \texttt{structured\_nofilter} (\S\ref{sec:mechanism}, a format-only ablation). Thus ``structured'' never means blindly dropping every non-current item; its retention rule depends on query intent.

Because \texttt{flat} ignores validity by construction, its BehavioralAcc is identical across all resolver tiers -- intentional, not an artifact: any behavioral shortfall under oracle-resolved, flat-assembled context is attributable entirely to assembly, since resolution is by definition perfect there. This is not a deployment claim; it isolates the upper bound of assembly's utility when validity labels are trustworthy, which we then relax in \S\ref{sec:reliability}--\S\ref{sec:mab}.

\section{ValidityMem-530}
\label{sec:dataset}

We construct a synthetic benchmark, \textsc{ValidityMem-530}, to obtain gold validity labels for controlled analysis. We avoid the name FreshBench because it already denotes a NAACL 2025 temporal-generalization benchmark \citep{zhu2025freshbench}; the released Python package retains its legacy \texttt{freshbench\_lite} directory name only for code compatibility. The benchmark comprises 530 queries over 1{,}936 unique stored memory objects across six scenario types: \texttt{current} (no conflict), \texttt{superseded\_monotonic} (a temporally monotonic update chain), \texttt{expired\_interval} (a closed validity window), \texttt{historical\_only} (paired present- and past-scoped questions over historical evidence), \texttt{non\_monotonic\_reversion} (a value is stated, changed, then reverted), and \texttt{implicit\_conflict} (conflicting statements with no explicit correction language).

\textbf{Example.} A \texttt{superseded\_monotonic} chain: \textit{Jamie: ``I work remote.''} then \textit{Jamie: ``I just switched to working in-office.''} -- gold answer \textit{in-office}, first statement labeled \texttt{superseded}. \texttt{implicit\_conflict} instances instead pair an explicit statement with a later, behaviorally inconsistent one and no correction language (e.g.\ a stated vegetarian later mentioning eating meat), probed with a yes/no conflict question.

\textbf{Construction and statistics.} Memories and queries are generated from a templated vocabulary of entities, slot values, and phrasing variants, with timestamps in varying styles. Each query references 9--11 candidates (mean 10.31): a relevant chain plus seeded, interleaved distractors. The 1{,}936 figure counts unique stored objects; reuse across paired queries yields 5{,}462 query--memory candidate occurrences. Scenario and question type are different axes: \texttt{superseded\_monotonic}, for example, contributes extraction, yes/no, and aggregation queries. Their exact cross-tabulation is: current $40$ extraction; superseded $60$ extraction, $60$ yes/no, $34$ aggregation; expired $80$ extraction, $40$ yes/no; historical-only $40$ present-state extraction and $40$ historical; reversion $50$ extraction and $50$ yes/no; and implicit conflict $36$ probes. These sum to extraction 270, yes/no 150, historical 40, probe 36, and aggregation 34. We release the generator and consistency checks; \textsc{ValidityMem-530} is a controlled diagnostic, not a replacement for naturalistic evaluation.

\section{Experimental Setup}
\label{sec:setup}

Experiments use GPT-4o-mini (OpenAI), Claude Haiku 4.5 (Anthropic), and a larger Claude Sonnet model. The two smaller models run at temperature 0; Sonnet rejects fixed-temperature sampling, and we disable its default adaptive thinking to match the direct-answer setup. Generation uses a 60-token cap. The same generation and adjudication prompts are used across models. The 6k MemoryAgentBench extraction prompt was iteratively refined after inspecting GPT-4o-mini failures on that sample, so those results are explicitly a tuned diagnostic; the subsequent question-disjoint 32k run freezes both extraction and generation prompts (\S\ref{sec:mab}). Hosted APIs are not exactly deterministic across time. We report paired, query-level bootstrap confidence intervals (2{,}000 resamples; fixed seed). All stored responses were regraded after freezing an abstention-aware committed-answer parser; a new blinded audit agrees on 118/120 responses (98.3\%, Wilson 95\% CI [94.1, 99.5], $\kappa=0.962$; Appendix~\ref{app:grading}).

\section{Resolver-Error Sensitivity Sweep}
\label{sec:reliability}

The oracle table uses all 530 queries. To control API cost, every corruption point uses a fixed $n=204$ subset: the first 34 queries in frozen file order from each of six scenario types, generated with benchmark seed 0 and selected before model calls. File order was not randomized, so we call this a scenario-balanced sensitivity subset, not a representative sample; Appendix~\ref{app:sensitivity-composition} reports its question-type composition. The original multi-model intervention uses corruption base seed 0; the matched three-policy rerun uses portable SHA-256-derived per-query seeds. Starting from gold labels, we independently replace each label with a uniformly random different state at $p\in\{0,.1,.2,.3,.4,.5\}$ and report realized ResolutionAcc. This deliberately broad sensitivity intervention is not a generative model of realistic resolver errors; targeted false-drop and false-keep interventions separately probe error structure. A cached \texttt{flat} answer per query removes baseline sampling noise.

\begin{figure*}[t]
\centering
\resizebox{\textwidth}{!}{\input{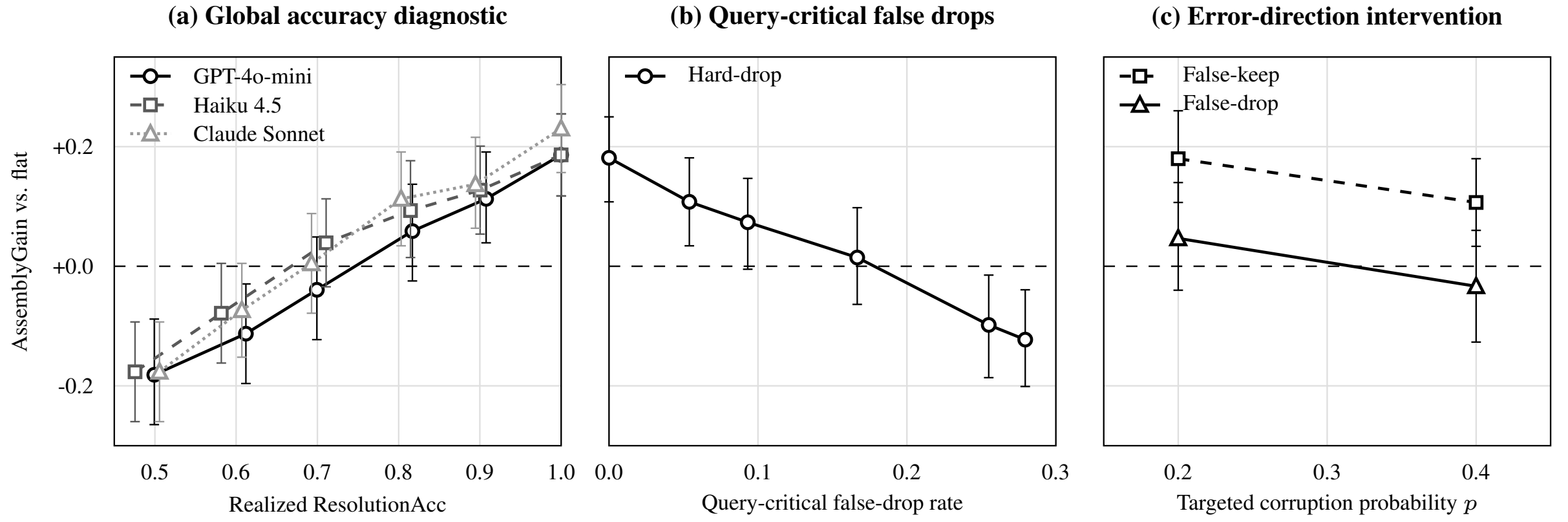}}
\caption{The mechanism is not captured by global accuracy alone. (a) Hard-drop gain versus gold-dependent ResolutionAcc ($n=204$ per point). (b) On the matched GPT-4o-mini sweep, gain declines with query-critical false-drop rate ($n=204$). (c) Targeted Haiku corruption shows false-drop is costlier than false-keep ($n=150$). Curves use distinct markers and line styles; error bars are paired query-bootstrap 95\% CIs and the horizontal dashed line is zero.}
\label{fig:reliability}
\end{figure*}

At perfect resolution, hard-drop gains are $+0.186$, $+0.186$, and $+0.230$ for GPT-4o-mini, Haiku, and Sonnet; at the lowest tested reliability they are $-0.181$, $-0.176$, and $-0.176$. AssemblyGain decreases monotonically as reliability degrades, with all three curves crossing zero. Linear interpolation places the diagnostic crossings near ResAcc $0.758$, $0.668$, and $0.688$, respectively, but these are model- and task-specific summaries rather than deployment thresholds. Appendix Table~\ref{tab:fullgrid} further shows why: heterogeneous resolvers with equal or higher global accuracy do not necessarily yield the gain implied by the randomized curves. Error structure, especially query-critical false drops, is the more operational concern.

\textbf{Non-deleting validity awareness is not automatically safe.} We ran the complete matched three-policy sweep on GPT-4o-mini: label-free flat, soft-retain (retain all evidence with validity labels and ordering), and hard-drop. Soft-retain keeps current-fact recall at $1.0$ throughout, whereas hard-drop recall falls to $0.721$ as query-critical false-drop rises to $0.279$. Nevertheless, soft-retain's gain declines from $+0.157$ [0.093, 0.216] to $-0.118$ [$-0.196$, $-0.044$]. At ResAcc $0.612$, both soft-retain ($-0.118$) and hard-drop ($-0.098$) significantly underperform flat. Preserving content prevents deletion loss but not misleading-label steering; when validity guidance is unvalidated, label-free flat is safer.

\textbf{False-drop is costlier than false-keep.} Targeted Haiku corruption isolates error direction using one shared, regraded flat baseline. At $p=.2$, Gain is $+0.180$ [0.107, 0.260] for false-keep versus $+0.047$ [$-0.040$, 0.140] for false-drop; the paired false-keep-minus-false-drop difference is $+0.133$ [0.073, 0.193]. At $p=.4$, the gains are $+0.107$ [0.033, 0.180] versus $-0.033$ [$-0.127$, 0.060], with paired difference $+0.140$ [0.053, 0.227]. Both paired CIs exclude zero ($n=150$): false-drop causes significantly greater downstream degradation than false-keep at both intervention levels. In the matched OpenAI sweep, hard-drop gain declines as query-critical false-drop rises from $0$ to $0.279$ (Figure~\ref{fig:reliability}b).

\textbf{A deployment-matched validation gate avoids most regret.} We treat each corruption level as a separate deployment environment and split the 204 queries once into 102 calibration and 102 held-out test queries, stratified 17/17 per scenario by a frozen hash. For each environment, the gate selects hard-drop only if its calibration accuracy exceeds flat, otherwise flat; choices are frozen before test evaluation. It selects hard-drop for $p=0,.1,.2$ and flat for $p=.3,.4,.5$. Pooled across 612 held-out environment--query outcomes, accuracy is 0.655 versus 0.569 always-flat, 0.641 always-hard-drop, and 0.583 always-soft-retain. A test-informed oracle policy reaches 0.662, leaving gate regret 0.007 versus 0.093, 0.021, and 0.078 for the three fixed policies. This is retrospective evidence on one matched sweep, not a universal gate; the split, choices, per-environment results, and scenario breakdown are frozen in Appendix~\ref{app:deployment-gate}.

\section{Oracle Upper Bound}
\label{sec:headline}

\begin{table}[!htb]
\centering
\scriptsize
\setlength{\tabcolsep}{2pt}
\begin{tabular}{lccc}
\toprule
Model & Flat & Hard-drop & AssemblyGain \\
\midrule
GPT-4o-mini & 0.600 & 0.775 & +0.175 [+0.130, +0.221] \\
Claude Haiku 4.5 & 0.558 & 0.760 & +0.202 [+0.157, +0.247] \\
Claude Sonnet & 0.583 & 0.817 & +0.234 [+0.185, +0.283] \\
\bottomrule
\end{tabular}
\caption{Oracle resolver on \textsc{ValidityMem-530}, $n=530$, regraded with the frozen committed-answer parser. All paired 95\% CIs exclude zero.}
\label{tab:headline}
\end{table}

With gold labels, hard-drop consistently improves behavior, establishing the high-reliability upper bound. The oracle result is useful setup rather than the main deployment claim: it shows that valid labels have limited value when the generator receives unresolved contradictions, while Figure~\ref{fig:reliability} identifies when trusting those labels becomes unsafe.

\section{Format versus Filtering}
\label{sec:mechanism}

\texttt{structured\_nofilter} applies the same headers and bullets to every memory but drops nothing. On GPT-4o-mini, format-only Gain is $+0.006$ [$-0.019$, $+0.028$] and filtering beyond that adds $+0.170$ [0.126, 0.215]. On Haiku, the corresponding values are $-0.023$ [$-0.051$, $+0.004$] and $+0.225$ [0.177, 0.270]. Formatting CIs include zero; the oracle benefit comes from filtering.

\FloatBarrier
\section{Published-Benchmark Diagnostic: MemoryAgentBench}
\label{sec:mab}

\begin{table*}[t]
\centering
\footnotesize
\begin{tabular}{llcccc}
\toprule
Model & Hop & Extraction architecture & Flat & Struct. & AssemblyGain (CI) \\
\midrule
GPT-4o-mini & single & one-shot & 0.360 & 0.800 & +0.440 [+0.330, +0.550] \\
GPT-4o-mini & multi & one-shot & 0.090 & 0.040 & $-$0.050 [$-$0.100, $-$0.010] \\
GPT-4o-mini & single & hop-decomp. & 0.370 & 0.780 & +0.410 [+0.300, +0.510] \\
GPT-4o-mini & multi & hop-decomp. & 0.140 & 0.280 & +0.140 [+0.070, +0.210] \\
Claude Haiku 4.5 & single & hop-decomp. & 0.400 & 0.680 & +0.280 [+0.190, +0.380] \\
Claude Haiku 4.5 & multi & hop-decomp. & 0.090 & 0.220 & +0.130 [+0.060, +0.210] \\
\bottomrule
\end{tabular}
\caption{Tuned diagnostic on the published MemoryAgentBench FactConsolidation benchmark, $n=100$ per row. The extraction prompt was refined after inspecting GPT-4o-mini failures on this evaluated sample; these are not held-out external-validation estimates.}
\label{tab:mab}
\end{table*}

\textsc{ValidityMem-530} is synthetic and self-authored. We therefore examine the same mechanism on MemoryAgentBench's published FactConsolidation task \citep{mab2025}, which tests selective forgetting through counterfactual edits to numbered facts. A higher serial number indicates a later restatement, making max-serial resolution correct \emph{conditional on accurate extraction}. We first use its 6k-token single- and multi-hop splits. Crucially, the extraction prompt was iteratively refined after diagnosing GPT-4o-mini failures on this same evaluated sample. We therefore label Table~\ref{tab:mab} a \emph{tuned diagnostic}, not held-out validation.

\textbf{Single-hop shows the predicted positive pattern.} Extraction is essentially perfect here (99/100 GPT-4o-mini questions extract exactly the one group a single-hop question needs), and hard-drop wins by a wide margin on both models. Because prompt development touched the evaluated benchmark, we treat even this clean mechanism slice as diagnostic rather than confirmatory.

\textbf{Multi-hop initially reversed the effect; we show why with a targeted intervention, not just observation.} With one-shot extraction, only 34/100 GPT-4o-mini multi-hop questions extracted the two fact-groups needed; most collapsed distinct relations into one group. When extraction merges facts from different relations, max-serial marks all but the highest-numbered member ``superseded,'' and structured assembly \emph{deletes} them -- even when a deleted fact belonged to a different, unresolved hop. Structured was worse than flat here (one-shot row, Table~\ref{tab:mab}): dropping content on an unreliable judgment is riskier than keeping everything.

We diagnosed this mechanism, hypothesized that fixing extraction would remove the reversal, and tested it: hop-decomposition extraction (decompose into sub-questions, extract per-hop, carry the prior hop's fact forward as a hint), re-ran, and the sign flipped as predicted (hop-decomp.\ row). This is interventional evidence -- diagnose, modify, confirm -- not a purely correlational observation; we do not claim a fully randomized manipulation here (\S\ref{sec:reliability} provides that). The same hop-decomp.\ procedure is applied unchanged to every hop-decomp.\ condition in Table~\ref{tab:mab}.

\textbf{Frozen, question-disjoint replication is inconclusive.} After freezing both prompts, we sampled 50 single-hop and 50 multi-hop questions by seeded SHA-256 rank from the previously untouched 32K-token split, excluding every normalized question duplicated in the tuned 6K-token sample. On GPT-4o-mini, answer-bearing extraction recall is 0.26 and query-critical false-drop incidence is 0.01; flat accuracy is 0.20 and hard-drop is 0.23, a paired Gain of $+0.03$ [$-0.01$, $+0.08$]. Extraction recall measures whether required answer evidence entered the resolver, whereas false-drop measures whether evidence reaching assembly was then deleted; upstream extraction misses are therefore reported separately. The 0.01 is the all-query incidence (1/100); conditional on answer evidence being extracted it is 1/26 (0.038). Multi-hop extraction recall is only 0.12. The direction is consistent with the tuned result but the interval includes zero, so this is a clean negative/inconclusive replication rather than external validation. Per-hop results and released provenance appear in Appendix~\ref{app:mab-heldout}.

\textbf{This is mechanistic corroboration, not independent replication.} With the corrected grader, Claude's multi-hop Gain is $+0.130$, close to GPT-4o-mini's $+0.140$, despite 22\% zero-extraction and under-extraction (mean 1.64 groups versus GPT-4o-mini's 4.06). The shared prompt was tuned on GPT-4o-mini and Claude lacks a one-shot row, so cross-model differences remain confounded. The randomized corruption sweep, not this diagnostic, is the load-bearing test of the conditional-reliability claim.

\FloatBarrier
\section{Discussion}
\label{sec:discussion}

Three probes converge: a token-budget selector drops the current fact under scarcity (Appendix~\ref{sec:budget}); label corruption drives AssemblyGain through zero, especially under false-drop (\S\ref{sec:reliability}); and unreliable extraction deletes facts that were never stale (\S\ref{sec:mab}). Hard-drop converts recoverable ambiguity into unrecoverable absence. Soft-retain removes that deletion channel but can still harm behavior by presenting incorrect validity labels as guidance. Systems should therefore treat deletion as a cost-sensitive decision: hard-drop only after validating query-critical false-drop risk on deployment-matched data. When such validation is unavailable, label-free retention is the safer default. Observed ResolutionAcc crossovers are model- and task-specific diagnostics, not universal deployment thresholds.

\section{Conclusion}

Validity resolution and downstream behavior are not the same quantity. Hard-drop improves accuracy with oracle labels but reverses sign as resolver errors accumulate. The asymmetric intervention identifies the mechanism: query-critical false drops are costlier than false keeps, while soft retention can still fail through misleading-label steering. A simple deployment-matched calibration gate recovers most test-informed oracle accuracy, but safe deployment still requires matched validation of false-drop risk rather than a universal threshold on global resolver accuracy.

\section{Limitations}
\label{sec:limitations}

\textbf{Three models from two providers, not a broad sweep.} GPT-4o-mini and Claude Haiku 4.5 are the two smaller models tested; the larger Claude Sonnet model shows the largest effect, but other frontier-scale models remain untested. The matched soft-retention sweep was conducted only on GPT-4o-mini, so cross-provider replication remains future work.

\textbf{Synthetic benchmark for the richer taxonomy.} \texttt{expired\_interval}, \texttt{historical\_only}, and \texttt{non\_monotonic\_reversion} are \textsc{ValidityMem-530}-only. The tuned MemoryAgentBench diagnostic covers the simpler current-vs-superseded case. Its frozen question-disjoint 32K-token replication is inconclusive and currently limited to GPT-4o-mini; independent cross-provider confirmation is still required.

\textbf{Only one end-to-end memory-system sanity check.} \S\ref{sec:mem0sanity} reports a small Mem0 baseline; we do not re-implement MemGPT or A-Mem. Our claim concerns a variable these systems do not isolate, not end-to-end system outperformance.

\textbf{Validity is not query relevance.} Our harness retains historical facts only when the question type itself is historical; a more general policy conditioning what to drop on query intent, not just validity labels, is untested here.

\textbf{Automatic grading retains a small false-positive mode.} The blinded audit reported in \S\ref{sec:setup} found two automated passes rejected by the annotator. Because its packet is balanced rather than prevalence-weighted, subgroup rates are diagnostic.

\textbf{Moderate sample sizes.} $n=530$ (\textsc{ValidityMem-530}) and $n=100$--$204$ per condition support the reported CIs; category-level and corruption splits are noisier.

\textbf{The deployment gate is retrospective.} Its split and choices are frozen, but it evaluates one model under controlled corruption regimes with labeled calibration outcomes. Prospective validation under naturally occurring resolver errors remains necessary.

\textbf{The multi-hop extraction fix is imperfect.} Hop-decomposition fixed under-extraction but overshot into over-extraction (mean 4.06 groups vs.\ target $\sim$2, GPT-4o-mini); the cross-model prompt-tuning confound is discussed in \S\ref{sec:mab}. Nor is the calibrated policy an unlabeled confidence gate: its behavioral validation outcomes require deployment-matched labels, while global ResolutionAcc does not capture error cost.

\section{Ethical Considerations}

This work uses synthetic data (\textsc{ValidityMem-530}, templated, no real personal data) and a published benchmark (MemoryAgentBench); we collect, use, or release no data involving real individuals. We see no risks beyond the general dual-use considerations applicable to any work on LLM agent memory.
In deployed memory systems, incorrect retention may preserve outdated or sensitive user information, while premature deletion may remove context needed for safe personalization; both operations should therefore be auditable and reversible where possible.

\bibliography{references}

\appendix

\section{Budget-Constrained Assembly}
\label{sec:budget}

Real deployments cannot give a resolver unlimited context budget. We sweep \texttt{structured\_budget}'s word budget over $\{15, 20, 30, 40, 60, 90, 150\}$ (oracle resolver, GPT-4o-mini, $n=102$). Regraded subset baselines are flat $=0.598$ and unconstrained hard-drop $=0.804$.

\begin{table}[ht]
\centering
\small
\begin{tabular}{lccc}
\toprule
Budget & Acc.\ & Curr.\ recall & Avg.\ words \\
\midrule
15 & 0.137 & 0.157 & 14.7 \\
20 & 0.255 & 0.304 & 20.3 \\
30 & 0.333 & 0.402 & 30.2 \\
40 & 0.480 & 0.520 & 40.2 \\
60 & 0.745 & 0.824 & 57.1 \\
90 & 0.804 & 0.833 & 59.0 \\
150 & 0.804 & 0.833 & 59.0 \\
\bottomrule
\end{tabular}
\caption{Budget sweep, oracle resolver, \texttt{structured\_budget}, $n=102$ (same subsample as the subset baselines above).}
\label{tab:budget}
\end{table}

The curve is monotonic, crosses flat between 40 and 60 words, and reaches the unconstrained 0.804 ceiling at 90 words. Accuracy closely tracks current-fact recall. The failure mode is therefore query-critical evidence loss under scarcity, parallel to false-drop under label corruption.

\section{Soft-Retention Mechanism Metrics}

\begin{table}[ht]
\centering
\scriptsize
\setlength{\tabcolsep}{2.5pt}
\begin{tabular}{lccccc}
\toprule
$p$ & ResAcc & NC P/R & Soft gain & Hard gain & Hard CR/FD \\
\midrule
0.0 & 1.000 & 1.000/1.000 & +0.157 & +0.181 & 1.000/0.000 \\
0.1 & 0.894 & 0.950/0.925 & +0.093 & +0.108 & 0.946/0.054 \\
0.2 & 0.832 & 0.916/0.916 & +0.054 & +0.074 & 0.907/0.093 \\
0.3 & 0.737 & 0.846/0.827 & $-$0.020 & +0.015 & 0.833/0.167 \\
0.4 & 0.612 & 0.769/0.796 & $-$0.118 & $-$0.098 & 0.745/0.255 \\
0.5 & 0.547 & 0.722/0.677 & $-$0.118 & $-$0.123 & 0.721/0.279 \\
\bottomrule
\end{tabular}
\caption{Matched GPT-4o-mini sweep ($n=204$ per point). NC P/R is micro non-current precision/recall; gains are versus cached flat accuracy 0.623; Hard CR/FD is current-fact recall/query-critical false-drop rate. Soft-retain CR is 1.0 and FD is 0 throughout.}
\label{tab:softmetrics}
\end{table}

\section{Sensitivity-Subset Composition}
\label{app:sensitivity-composition}

Every scenario contributes exactly 34 queries to the $n=204$ corruption subset. Because selection used the first queries in frozen, non-randomized file order, the subset is scenario-balanced rather than representative. It covers all five question types but overrepresents conflict probes and underrepresents yes/no and aggregation questions relative to the full benchmark.

\begin{table}[ht]
\centering
\scriptsize
\setlength{\tabcolsep}{3pt}
\begin{tabular}{lrrrr}
\toprule
Question type & Subset $n$ & Subset \% & Full $n$ & Full \% \\
\midrule
extraction & 105 & 51.5 & 270 & 50.9 \\
yes/no & 41 & 20.1 & 150 & 28.3 \\
historical & 17 & 8.3 & 40 & 7.5 \\
probe & 34 & 16.7 & 36 & 6.8 \\
aggregation & 7 & 3.4 & 34 & 6.4 \\
\bottomrule
\end{tabular}
\caption{Question-type composition of the scenario-balanced sensitivity subset versus \textsc{ValidityMem-530}.}
\label{tab:sensitivity-composition}
\end{table}

\section{Deployment-Gated Policy Evaluation}
\label{app:deployment-gate}

The same scenario-stratified split is reused across all six corruption regimes, preventing query overlap between calibration and test. Within each regime, the gate uses only calibration correctness labels. The oracle is a descriptive test-informed upper bound over flat, hard-drop, and soft-retain; it selects the same actions as an oracle restricted to flat versus hard-drop.

\begin{table}[ht]
\centering
\scriptsize
\setlength{\tabcolsep}{3pt}
\begin{tabular}{lrr}
\toprule
Policy & Test accuracy & Regret \\
\midrule
Always flat & 0.569 & 0.093 \\
Always hard-drop & 0.641 & 0.021 \\
Always soft-retain & 0.583 & 0.078 \\
Calibration-selected & \textbf{0.655} & \textbf{0.007} \\
Oracle selection & 0.662 & 0.000 \\
\bottomrule
\end{tabular}
\caption{Pooled performance over 612 held-out regime--query outcomes. Regret is accuracy loss relative to oracle policy selection.}
\label{tab:deployment-gate-policies}
\end{table}

\begin{table}[ht]
\centering
\scriptsize
\setlength{\tabcolsep}{2.5pt}
\begin{tabular}{rrrrll}
\toprule
$p$ & Cal. gain & Test flat & Test hard & Gate & Oracle \\
\midrule
0.0 & +0.137 & 0.569 & 0.794 & hard & hard \\
0.1 & +0.059 & 0.569 & 0.725 & hard & hard \\
0.2 & +0.010 & 0.569 & 0.706 & hard & hard \\
0.3 & $-$0.010 & 0.569 & 0.608 & flat & hard \\
0.4 & $-$0.176 & 0.569 & 0.549 & flat & flat \\
0.5 & $-$0.137 & 0.569 & 0.461 & flat & flat \\
\bottomrule
\end{tabular}
\caption{Frozen decisions by corruption regime. Cal. gain is calibration hard-drop minus flat accuracy; ties and negative values select flat.}
\label{tab:deployment-gate-regimes}
\end{table}

\section{Scenario-Level Sensitivity}
\label{app:scenario-sensitivity}

At perfect labels, hard-drop is non-harmful in all six scenarios and helps most on non-monotonic reversion. At the largest corruption level, it becomes harmful in five scenarios; reversion remains positive. Thus scenario type alone is not a safety certificate: the same monotonic-supersession category changes from $+0.059$ to $-0.235$ as resolver errors increase.

\begin{table}[ht]
\centering
\scriptsize
\setlength{\tabcolsep}{2.2pt}
\begin{tabular}{lrrrrr}
\toprule
Scenario & Flat & Hard$_{p=0}$ & Gain & Hard$_{p=.5}$ & Gain \\
\midrule
current & 1.000 & 1.000 & 0.000 & 0.706 & $-$0.294 \\
expired & 0.382 & 0.647 & +0.265 & 0.324 & $-$0.059 \\
historical & 0.882 & 0.971 & +0.088 & 0.588 & $-$0.294 \\
implicit conflict & 0.294 & 0.559 & +0.265 & 0.265 & $-$0.029 \\
reversion & 0.471 & 0.882 & +0.412 & 0.647 & +0.176 \\
superseded & 0.706 & 0.765 & +0.059 & 0.471 & $-$0.235 \\
\bottomrule
\end{tabular}
\caption{Matched GPT-4o-mini sensitivity subset ($n=34$ per scenario). Flat is invariant across corruption; hard-drop is shown at perfect labels and the largest corruption intervention.}
\label{tab:scenario-sensitivity}
\end{table}

\section{Held-Out MemoryAgentBench Detail}
\label{app:mab-heldout}

\begin{center}
\begin{minipage}{\columnwidth}
\centering
\scriptsize
\setlength{\tabcolsep}{2.5pt}
\begin{tabular}{lrrrrr}
\toprule
Hop & $n$ & Extract & FD & Flat/Hard & Gain [95\% CI] \\
\midrule
single & 50 & 0.40 & 0.00 & 0.32/0.38 & +0.06 [$-$0.02, 0.14] \\
multi & 50 & 0.12 & 0.02 & 0.08/0.08 & 0.00 [0.00, 0.00] \\
overall & 100 & 0.26 & 0.01 & 0.20/0.23 & +0.03 [$-$0.01, 0.08] \\
\bottomrule
\end{tabular}
\captionof{table}{Frozen question-disjoint 32K-token GPT-4o-mini replication. Extract is answer-bearing extraction recall; FD is all-query query-critical false-drop incidence. Conditional FD among extracted-evidence cases is 0/20 (single), 1/6 (multi), and 1/26 (overall). Gain CIs use paired query bootstrap.}
\label{tab:mab-heldout}
\end{minipage}
\end{center}

\section{Resolver Reliability Tiers}
\label{sec:tiers}

A discrete, 4-tier complement to \S\ref{sec:reliability}'s controlled sensitivity intervention.

\textbf{What \texttt{deterministic} and \texttt{hybrid} actually do.} \texttt{deterministic} orders same-(entity, slot) memories chronologically from surface time expressions and exact-text dedup, then infers interval expiry from expiry phrases -- no LLM calls, the harness's analogue of a max-timestamp resolution recipe. Its main failure is mechanical: exact-text dedup treats a literal restatement as a no-op, so a \texttt{non\_monotonic\_reversion} chain (A, then B, then A again) collapses the final reassertion into the first mention. \texttt{hybrid} escalates any chain the dedup step flags as ambiguous to an LLM adjudicator, which sees the un-deduplicated candidates and names which is current -- exactly the case designed to defeat \texttt{deterministic}'s shortcut (prompts in Appendix~\ref{app:prompts}).

\begin{table}[!htb]
\centering
\scriptsize
\renewcommand{\arraystretch}{0.88}
\begin{tabular}{lccc}
\toprule
Resolver & ResAcc & Flat/Struct.\ BehAcc & Gain \\
\midrule
\multicolumn{4}{l}{\textit{GPT-4o-mini}} \\
no\_labels & 0.474 & 0.600 / 0.600 & 0.000 \\
determ. & 0.812 & 0.600 / 0.600 & 0.000 \\
hybrid & 0.799 & 0.600 / 0.609 & +0.009 \\
oracle & 1.000 & 0.600 / 0.774 & +0.174 \\
\midrule
\multicolumn{4}{l}{\textit{Claude Haiku 4.5}} \\
no\_labels & 0.474 & 0.557 / 0.570 & +0.013 \\
determ. & 0.812 & 0.557 / 0.570 & +0.013 \\
hybrid & 0.857 & 0.557 / 0.653 & +0.096 \\
oracle & 1.000 & 0.557 / 0.758 & +0.202 \\
\midrule
\multicolumn{4}{l}{\textit{Claude Sonnet}} \\
no\_labels & 0.474 & 0.587 / 0.594 & +0.008 \\
determ. & 0.812 & 0.587 / 0.613 & +0.026 \\
hybrid & 0.897 & 0.587 / 0.728 & +0.142 \\
oracle & 1.000 & 0.587 / 0.823 & +0.236 \\
\bottomrule
\end{tabular}
\caption{Regraded AssemblyGain across heterogeneous resolver tiers. Oracle is largest for every model, but global accuracy does not form a calibrated deployment axis.}
\label{tab:fullgrid}
\end{table}

Table~\ref{tab:fullgrid} shows the same qualitative ordering without treating heterogeneous resolver tiers as a calibrated axis. GPT-4o-mini's non-oracle gains are $0$--$0.009$; Haiku's rise from $+0.013$ to $+0.096$ as the tier improves; and Sonnet's higher-reliability hybrid reaches $+0.142$. Oracle is largest for every model. The controlled corruption sweep in Figure~\ref{fig:reliability}, rather than this heterogeneous grid, supports the monotonic claim.

\section{Grading Robustness and Manual Audit}
\label{app:grading}

The first grader used raw substring matching; a second used token boundaries but still accepted a gold value mentioned after a leading \texttt{UNKNOWN}. We now freeze a committed-answer grader with three explicit rules: (1) leading \texttt{UNKNOWN} or an equivalent abstention always fails; (2) yes/no is determined by the first committed yes/no token, never later substring occurrence; and (3) categorical surface variants are normalized through a released, task-defined alias map (e.g., \textit{Office} and \textit{in-office} for \textit{working from the office}). Regression tests cover abstention leakage, yes/no reversal, embedded numerals, punctuation, and aliases.

We regraded every stored output used in the manuscript without making new model calls. All three oracle headline CIs remain above zero. All three reliability curves still move from significantly positive at high reliability to significantly negative at low reliability and remain monotonic as reliability degrades. One near-zero Sonnet point changes sign ($-0.005\to+0.005$), shifting its crossover bracket by one grid interval; the reversal survives. MemoryAgentBench and budget numbers were also recomputed. Full machine-readable pre/post provenance is released in \texttt{regraded\_results.json}.

The earlier 80-item audit discovered the remaining bugs and therefore cannot validate the grader that those findings helped define. We replaced it with a new blinded 120-item packet: 40 responses per model, 20 per scenario, and 60 per assembly condition, with model, condition, scenario, and automated verdict hidden. The annotator completed all labels before the private key was opened. Agreement is 118/120 (98.3\%, Wilson 95\% CI [94.1, 99.5]), with Cohen's $\kappa=0.962$ and confusion counts TP=80, FP=2, FN=0, TN=38 when the human label is the reference. Agreement is 100\% for GPT-4o-mini, Sonnet, and hard-drop responses, 95.0\% for Haiku, and 96.7\% for flat responses. Both disagreements are flat Haiku outputs that mention the gold value without committing to it (an alternative answer and a future state). We do not tune the frozen grader on these audited cases; within the audit, both residual errors favor flat rather than hard-drop, so they do not suggest an inflated AssemblyGain. The completed packet and aggregate report are released.

\section{Prompts and Reproducibility Detail}
\label{app:prompts}
The exact model identifiers are \texttt{gpt-4o-mini}, \texttt{claude-haiku-4-5-20251001}, and \texttt{claude-sonnet-5}; per-experiment API call counts are provided in the supplementary code release. Below are the three system prompts underlying the paper's results, close to verbatim (minor reformatting for space); user-turn templates are abbreviated.

\subsection{Generation prompt}
Used by the Generator $G$ for every \textsc{ValidityMem-530} and MemoryAgentBench answer, regardless of assembly mode.

\textit{System:} ``You are answering a question using ONLY the memory context provided. Do not use outside knowledge. If the context does not contain the answer, reply exactly: UNKNOWN. Answer as briefly as possible -- a single value, name, number, or Yes/No -- with no explanation.''

\textit{User:} \texttt{Memory context: \{assembled context\}\ \ Question: \{query text\}\ \ Answer:}

\subsection{Hybrid adjudication prompt}
Used by the \texttt{hybrid} resolver tier (\S\ref{sec:tiers}) to escalate chains its deterministic dedup step flags as ambiguous. Sees only surface text and surface time expressions -- never the ground-truth timestamp.

\textit{System:} ``You are tracking which of several statements about the same fact is still true as of a given date. Read the statements in the order given (that is NOT necessarily chronological order). Reply with ONLY the number of the statement that is current as of the given date -- nothing else.''

\textit{User:} \texttt{As of \{query\_time\}: 1.\ \{statement\} (mentioned: \{surface\_time\_expr\}) 2.\ ...\ \ Which statement number is current?}

\subsection{MemoryAgentBench extraction prompt}
Used to extract candidate fact groups from MAB's numbered fact list (\S\ref{sec:mab}). The one-shot version is used directly; the hop-decomp.\ version first decomposes the question (a separate call, prompt below), then runs this same extraction prompt once per sub-question, carrying the previous hop's resolved fact forward as a hint.

\textit{Extraction system:} ``You extract facts relevant to a question from a numbered list of facts. Facts are numbered starting at 0. Some facts have been edited: a LATER (higher-numbered) fact about the exact same subject and relation OVERRIDES an earlier fact with the same subject and relation but a different value. Find every fact relevant to answering the question -- including BOTH the original and any edited/overriding versions of it -- and group them: each group is a list of facts about the exact same (subject, relation) pair. For multi-hop questions, include a separate group for EVERY hop needed to answer the question. Return ONLY strict JSON: a list of groups, each group a list of objects with ``number'' (int) and ``text'' (string, copied exactly from the fact list). No prose, no markdown fences, no explanation.''

\textit{Decomposition system (hop-decomp.\ only):} ``You decompose a multi-hop question into an ordered sequence of atomic single-hop sub-questions. Each sub-question after the first should be answerable only once the previous hop's answer is known. Return ONLY strict JSON: a list of strings, each one sub-question, in order.''

\section{ValidityMem-530 Scenario Summary}
\label{app:scenarios}

\begin{table}[ht]
\centering
\scriptsize
\setlength{\tabcolsep}{3pt}
\begin{tabular}{p{0.32\linewidth}p{0.4\linewidth}cc}
\toprule
Scenario & Example question & $n$ & Confl.? \\
\midrule
\texttt{current} & ``What is Priya's current city of residence?'' & 40 & No \\
\texttt{superseded\_}\newline\texttt{monotonic} & ``What is Jamie's current work mode?'' & 154 & Yes \\
\texttt{expired\_interval} & ``What is Hiro's work mode right now?'' & 120 & Yes \\
\texttt{historical\_only} & ``What was Sam's city of residence back in 2024?'' & 80 & Yes \\
\texttt{non\_monotonic\_}\newline\texttt{reversion} & ``What is Jordan's current employer?'' & 100 & Yes \\
\texttt{implicit\_}\newline\texttt{conflict} & ``Would Tariq be comfortable ordering a steak dinner tonight?'' & 36 & Yes \\
\bottomrule
\end{tabular}
\caption{\textsc{ValidityMem-530} scenario types, one generated example per type, and query count. Historical-only has 40 present-state extraction and 40 historically scoped queries; the example shows the latter.}
\label{tab:scenarios}
\end{table}

\end{document}